\documentclass[conference]{IEEEtran}
\IEEEoverridecommandlockouts
\usepackage{cite}
\usepackage{amsmath,amssymb,amsfonts}
\usepackage{algorithm}
\usepackage{algpseudocode}
\usepackage{graphicx}
\usepackage{textcomp}
\usepackage{xcolor}
\usepackage{rotating}
\usepackage{booktabs}
\def\BibTeX{{\rm B\kern-.05em{\sc i\kern-.025em b}\kern-.08em
    T\kern-.1667em\lower.7ex\hbox{E}\kern-.125emX}}
\begin{document}

\title{HybridFL: A Federated Learning Approach for Financial Crime Detection}

%
\author{
    Afsana Khan\IEEEauthorrefmark{1}, 
    Marijn ten Thij\IEEEauthorrefmark{2}, 
    Guangzhi Tang\IEEEauthorrefmark{1}, 
    Anna Wilbik\IEEEauthorrefmark{1} \\
    \IEEEauthorrefmark{1}Department of Advanced Computing Sciences, Maastricht University, Netherlands \\
    \IEEEauthorrefmark{2}Department of Cognitive Science and Artificial Intelligence, Tilburg University, Netherlands \\
    Corresponding author: a.khan@maastrichtuniversity.nl
}

\maketitle

\begin{abstract}
Federated learning (FL) is a privacy-preserving machine learning paradigm that enables multiple parties to collaboratively train models on privately owned data without sharing raw information. While standard FL typically addresses either horizontal or vertical data partitions, many real-world scenarios exhibit a complex hybrid distribution. This paper proposes Hybrid Federated Learning (HybridFL) to address data split both horizontally across disjoint users and vertically across complementary feature sets. We evaluate HybridFL in a financial crime detection context, where a transaction party holds transaction-level attributes and multiple banks maintain private account-level features. By integrating horizontal aggregation and vertical feature fusion, the proposed architecture enables joint learning while strictly preserving data locality. Experiments on AMLSim and SWIFT datasets demonstrate that HybridFL significantly outperforms the transaction-only local model and achieves performance comparable to a centralized benchmark.
\end{abstract}


\section{Introduction}
Federated learning (FL) enables multiple parties to collaboratively train machine learning models without sharing raw data, keeping data local and exchanging model parameters, updates, or intermediate representations during training \cite{mcmahan2017communication,yang2019federated}. This approach is valuable when data are distributed across organizational silos and cannot be centralized due to regulatory, contractual, commercial, or operational constraints \cite{kairouz2021advances}. Instead of transferring individual records, data owners share model updates or intermediate representations, reducing privacy risks while retaining the benefits of collaborative learning. 

Federated learning is commonly studied under two data partition types \cite{zhao2018federated}. In horizontal federated learning, participants hold disjoint subsets of samples described by the same feature space. In vertical federated learning, participants hold different and complementary feature subsets describing the same or largely overlapping set of samples, and labels are held by one party depending on the task. While horizontal federated learning benefits from broader sample coverage, vertical federated learning enables feature fusion across parties. 

However, many real-world collaborations do not fit neatly into either structure. Sample space may overlap only partially across organizations, feature space may be fragmented among different holders, and labels may be available only to a subset of participants. For such hybrid data partitions, learning must combine horizontal aggregation and vertical fusion within a single process. Treating hybrid partitioned data as purely horizontal can ignore complementary feature information, while treating it as purely vertical can exclude samples outside the overlap. Both simplifications can reduce model utility. These limitations motivate hybrid federated learning, which combines horizontal collaboration among similar parties with vertical fusion across complementary roles while keeping raw data local.

This paper investigates hybrid partitions where some organisations hold disjoint subsets of users, representing horizontal relationships, and complementary features exist only for users shared across certain organisations, representing vertical relationships. To address this setting, we propose a hybrid federated learning method that integrates horizontal aggregation and vertical fusion within a unified training process. Our contributions are as follows:
\begin{itemize}
\item We design a hybrid federated learning algorithm capable of learning effectively from complex partitions, where data are split both horizontally across disjoint user groups and vertically across complementary feature sets.
\item We evaluate the proposed method using realistic synthetic datasets in the context of financial crime detection. The results demonstrate that the hybrid approach achieves performance comparable to centralised learning while maintaining data privacy.
\end{itemize}

\section{Related Work}
Horizontal FL and vertical FL differ in both what is shared during training and how predictions are produced at inference time. In horizontal FL \cite{liu2022towards}, all parties train the same model on their own local samples that share a common feature space. Training typically proceeds in rounds where a coordinating server distributes the current global parameters, clients run several local optimization steps, and the server aggregates client updates, commonly through weighted averaging as in FedAvg \cite{mcmahan2017communication}. Once training is complete, inference is usually local because each client has the full feature set required to make predictions on new data at its end, so no cross-party interaction is needed at inference.

In vertical FL \cite{khan2025vertical}, parties hold complementary feature subsets for the same or largely overlapping samples, so training and inference must both be coordinated across parties. One party, often called the active party, holds the labels and runs the prediction module, or top model. The remaining parties are passive parties that hold different feature subsets and maintain local models that transform their private features into intermediate representations. During training, passive parties compute these representations and share them with the active party. The active party computes the loss, derives gradients with respect to each shared representation, and sends only those gradients back to the corresponding passive parties so they can update their models end-to-end without revealing raw features. Because no single party has all the required features, inference in a vertical FL setting generally follows a coordinated protocol where the active party combines representations received from passive parties to perform the inference.

Financial fraud and risk modeling often require information that sits in separate institutions, while direct data sharing is limited by privacy, compliance, and operational restrictions. Federated learning provides a way to collaborate under these constraints without moving raw records. Much of the early studies, therefore, adopt a horizontal FL formulation, where multiple banks with the same feature schema collaboratively train a shared fraud model by aggregating locally computed updates, as in federated credit card fraud detection, where each bank keeps its transaction logs local but benefits from a broader effective training population \cite{yang2019ffd}. Vertical FL also appears in finance when different organizations hold complementary attributes about the same customers. In this setting, a label holder such as a lender acts as the active party and trains the prediction model, while other institutions act as passive parties that contribute additional customer features without revealing them \cite{cheng2021secureboost}.
These studies establish that FL can be effective for financial fraud and risk modeling, but many settings are still approximated as purely horizontal or vertical, which does not reflect how information is distributed in payment networks.

Interbank payment workflows naturally induce a hybrid partition. Transaction-level features and labels may be available at a transaction processing entity, while account-level histories and behavioral indicators remain within the sender and receiver banks. The learning sample is relational because each transaction links two accounts that may belong to different banks, and banks hold disjoint customer populations. This structure combines horizontal separation across banks with vertical feature dependency across the roles that participate in a transaction.

Recent work addresses this hybrid structure directly for financial crime detection. Zhang et al. proposed a privacy-preserving hybrid FL framework that combines transaction-side information with bank-side account activity and discusses privacy risks across both training and inference \cite{zhangprivacy}. Fed-RD formalizes hybrid learning for relational transaction data that is vertically and horizontally partitioned and studies privacy-preserving mechanisms and fusion design choices together with secure aggregation and differential privacy \cite{khan2024fed}. These papers motivate hybrid FL as a suitable approach for payment-network fraud settings while focusing either on system-level privacy considerations or on privacy mechanisms and analysis. In this paper, we focus on a practical end-to-end hybrid training architecture aligned with the sender–receiver structure of interbank payments. We maintain separate sender and receiver encoders at each bank because outgoing and incoming activity often reflects different behavioral patterns and risk signals, and a single shared encoder can blur role-specific cues. Training is coordinated through embedding-level gradient exchange with the banks involved in each transaction, and bank encoders are periodically synchronized through federated averaging to reduce bank-specific drift and keep embeddings aligned across institutions for the top model.

\section{Problem Statement}
In hybrid federated learning, data are distributed across multiple organisations both horizontally and vertically. Each organisation may observe only a subset of users and a subset of the features describing those users. This fragmented distribution means that no single participant holds a complete view of the data, either across entities or across attributes. Learning from such hybrid partitions is challenging because the relationships among users and features extend across different organisations and roles. Disjoint user groups prevent straightforward aggregation of gradients, while complementary feature spaces require joint optimisation without direct data sharing. The coexistence of overlapping and disjoint data segments complicates both coordination and privacy management. Conventional horizontal and vertical federated learning approaches address only one dimension of this problem and therefore cannot fully capture the dependencies present in hybrid data distributions. 

In this paper, we consider hybrid federated learning under a structured setting that enables coordinated model training across multiple parties holding complementary and partially overlapping data. 
Each party $m \in \{1, \dots, M\}$ belongs to a specific role defined by its feature space and holds local data
\[
\mathcal{D}_m = \{(x_i^{m}, y_i^{m})\}_{i=1}^{N_m},
\]
where $x_i^{m}$ denotes the features available to party $m$ and $y_i^{m}$ the corresponding labels, if present. 
Parties of the same type maintain disjoint subsets of users with similar feature spaces, representing horizontal partitioning, while parties of different types contribute complementary feature subsets that describe overlapping entities participating in shared samples, representing vertical partitioning. 
This structured hybrid arrangement allows related records across roles (for example, entities interacting in a common event) to be linked during collaborative training without any exchange of raw data. The learning objective is to train a global model $f(\cdot; \Theta)$, parameterised by $\Theta$, that minimises the expected loss over the joint data distribution $\mathcal{D}$ while ensuring data locality:
\[
\min_{\Theta} \; \mathbb{E}_{(x, y) \sim \mathcal{D}} \, \mathcal{L}\big(f(x; \Theta), y\big),
\]
subject to the constraint that each party $m$ can compute only on its own local data $\mathcal{D}_m$. 
Optimisation is achieved through coordinated local computation and controlled information exchange across horizontally aligned and vertically linked parties. The goal is to approximate the global minimiser of $\mathcal{L}$ under distributed ownership, structured overlap, and privacy constraints.

We instantiate this problem in the context of \emph{financial crime detection}, where the structured hybrid configuration naturally arises from the interactions among sender banks, receiver banks, and the transaction party. This use case provides a concrete and realistic setting in which hybrid federated learning addresses practical challenges of data fragmentation, privacy, and collaboration among financial institutions. The financial crime detection scenario considered in this paper involves collaboration among institutions that collectively participate in the processing of interbank transactions. The system consists of three main roles. Sender banks maintain information about the accounts initiating payments, including account identifiers, historical activities, and behavioural indicators. Receiver banks store comparable data for the accounts that receive funds. These institutions form the horizontally partitioned participants, as each bank holds data for a distinct set of customers. A third role, referred to as the transaction party, corresponds to the entity responsible for recording and managing the transaction flow between banks, such as a central payment processor, clearing house, or regulatory network operator like SWIFT. This party stores transaction-level attributes such as amount, currency, timestamp, and basic metadata, as well as the labels indicating whether each transaction is legitimate or suspicious.
\begin{figure*}[h]
\centering
\includegraphics[width=\textwidth]{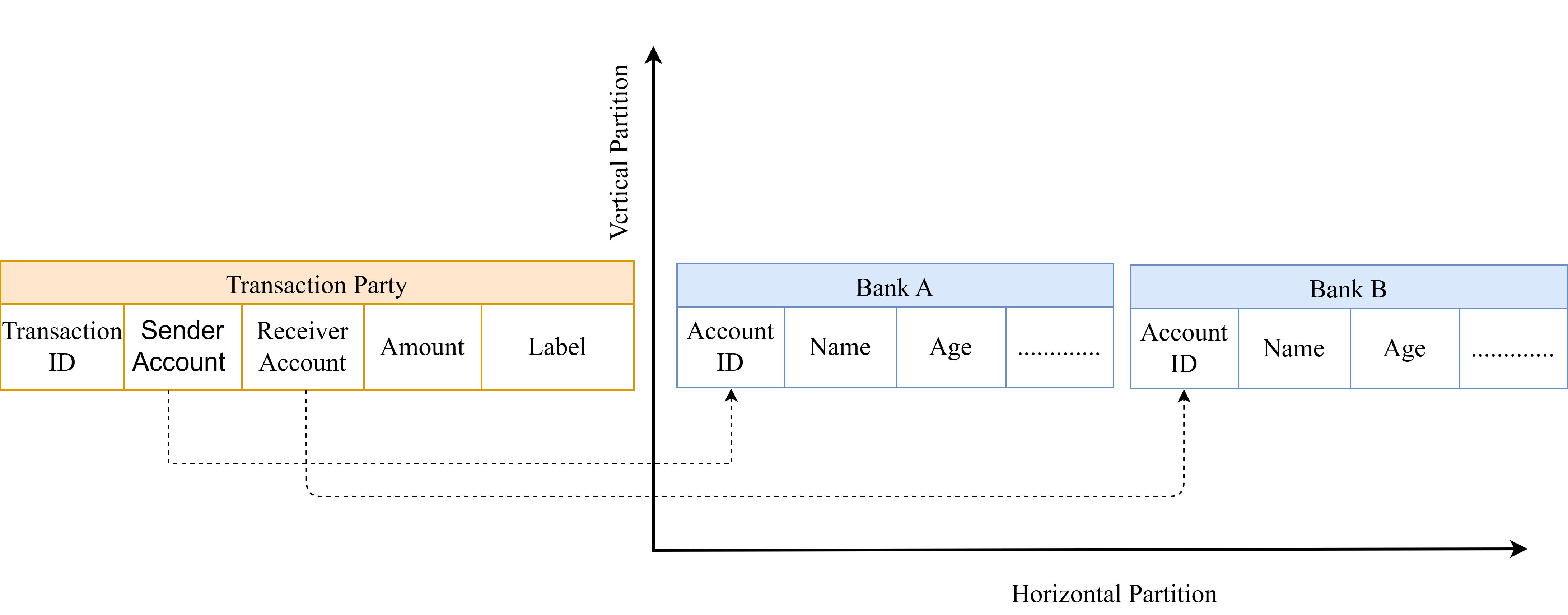}
\caption{Illustration of the hybrid data partition. The transaction party holds transaction-level records, while Banks A and B each maintain private account-level data for their respective customers. Transactions link the sender and receiver accounts, creating vertical connections across institutions and horizontal separation within each bank.}
\label{fig:hybrid_partition}
\end{figure*}
The overall architecture produces a hybrid data distribution as shown in Figure \ref{fig:hybrid_partition}. Horizontally, banks hold non-overlapping customer records, while vertically, transactions connect complementary information from multiple organisations. No single institution possesses a complete view of an event, since transaction details and account-level data are distributed across roles. The system therefore provides a realistic environment for hybrid federated learning, where collaboration between banks and the transaction party can enable joint analysis of financial activity without direct data sharing. 
\section{Proposed Method}

In the financial crime detection setup introduced earlier, multiple institutions participate in interbank transactions while holding different parts of the relevant information. The proposed hybrid HybridFL algorithm operates within this setting to enable joint model training without exchanging raw data. The system includes one \emph{active party}, referred to as the transaction party, and multiple \emph{passive parties}, corresponding to the participating banks. The transaction party coordinates the training process, holds the transaction-level features and labels, and maintains two models: (i) a \emph{local model} $h_T(\cdot;\theta_T)$ that computes a transaction embedding, and (ii) a \emph{top model} or fusion head $f_F(\cdot;\theta_F)$ that produces the final prediction.

Each interbank transaction involves two accounts that may belong to different banks and is recorded by the transaction party. The transaction party stores the transaction features $x_i^T$ (such as amount, time, currency, and channel) together with the label $y_i$. Each bank $b \in \mathcal{B}$ maintains account-level features for its own customers, denoted $x_i^{b}$, which are accessed only when that customer appears as sender or receiver. Because an account may initiate or receive payments, each bank maintains both a sender encoder and a receiver encoder, enabling separate representations for outgoing and incoming behaviour while keeping all computation local. The sender and receiver encoders are parameterised by $\theta_S^b$ and $\theta_R^b$ and compute
\[
e_i^{S,b} = h_S(x_i^{b};\theta_S^b), \qquad
e_i^{R,b} = h_R(x_i^{b};\theta_R^b),
\]
depending on whether bank $b$ is the sender or receiver for sample $i$. The transaction party computes a transaction-level embedding using its local model,
\[
e_i^T = h_T(x_i^T;\theta_T).
\]
The fusion head combines these embeddings to produce the final prediction through the top model,
\[
\hat{y}_i = f_F\!\left(\text{concat}(e_i^{T}, e_i^{S,b_S}, e_i^{R,b_R});\theta_F\right).
\]
For a minibatch $\mathcal{B}$, the loss is evaluated at the transaction party as
\[
\mathcal{L} = \frac{1}{|\mathcal{B}|}\sum_{i\in\mathcal{B}}
\mathcal{L}(y_i,\hat{y}_i).
\]
\begin{figure*}[h]
\centering
\includegraphics[width=\textwidth]{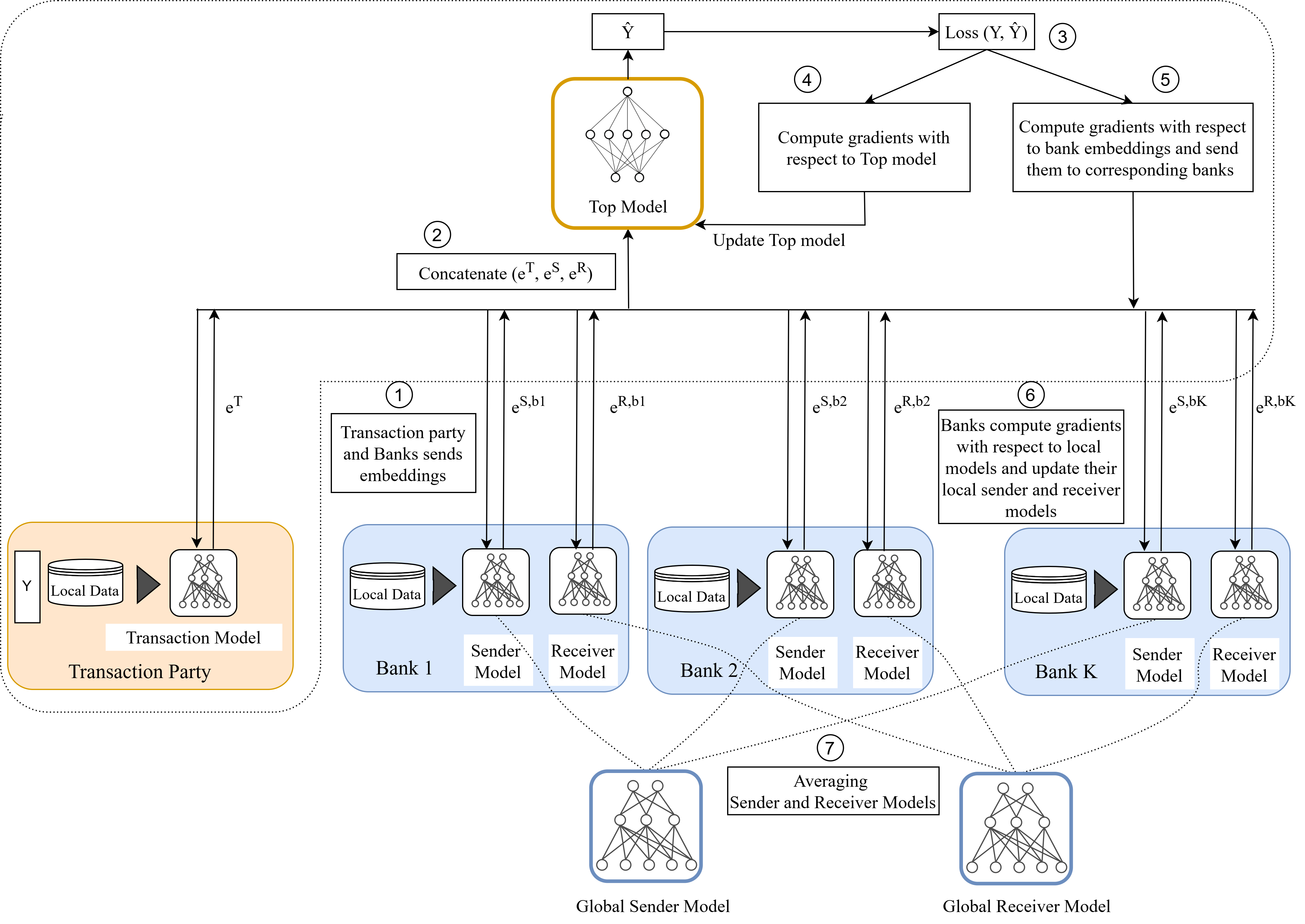}
\caption{HybridFL training workflow: The transaction party acts as the active party, fusing embeddings received from the sender and receiver banks (passive parties) with its own transaction embedding to produce a prediction and compute the loss and gradients with respect to embeddings. Computed gradients are sent back to the corresponding banks for updating local sender and receiver models. Bank local models are periodically synchronized via Federated Averaging}
\label{fig:hybrid_fl}
\end{figure*}

Training proceeds collaboratively. The transaction party computes the loss and its gradients with respect to the three embeddings $e_i^T$, $e_i^{S,b_S}$, and $e_i^{R,b_R}$. The gradients corresponding to sender and receiver embeddings are transmitted to the respective banks. Each bank then performs local backpropagation through its sender or receiver encoder to update $\theta_S^b$ and $\theta_R^b$. In parallel, the transaction party updates the transaction encoder and fusion head. Because all banks must learn compatible representations, their sender and receiver encoders are periodically synchronised using federated averaging (Figure~\ref{fig:hybrid_fl}).

\begin{algorithm}[h]
\footnotesize
\caption{Hybrid Federated Learning (HybridFL)}
\label{alg:hyfl}
\begin{algorithmic}[1]

\State \textbf{Active party:} transaction party with dataset $\mathcal{D}^T$, local model $h_T(\cdot;\theta_T)$, top model $f_F(\cdot;\theta_F)$
\State \textbf{Passive parties:} banks $b$ with datasets $\mathcal{D}^b$, sender encoder $h_S(\cdot;\theta_S^b)$, receiver encoder $h_R(\cdot;\theta_R^b)$

\For{each round}
    \State Sample minibatches $\{\mathcal{B}\}$ from $\mathcal{D}^T$
    \For{each minibatch $\mathcal{B}$}

        \For{$i\in\mathcal{B}$}
            \State $b_S,b_R \leftarrow$ sender and receiver banks for transaction $i$
            \State $e_i^{S,b_S} = h_S(x_i^{b_S};\theta_S^{b_S})$
            \State $e_i^{R,b_R} = h_R(x_i^{b_R};\theta_R^{b_R})$
            \State $e_i^T = h_T(x_i^T;\theta_T)$
            \State $\hat{y}_i = f_F(\text{concat}(e_i^T, e_i^{S,b_S}, e_i^{R,b_R});\theta_F)$
        \EndFor

        \State $\mathcal{L}
        = \frac{1}{|\mathcal{B}|}\sum_{i\in\mathcal{B}}
        \mathcal{L}(y_i,\hat{y}_i)$

        \State $\nabla_{\theta_F} = \frac{\partial\mathcal{L}}{\partial\theta_F}$
        \State $\nabla_{e_i^T},\;\nabla_{e_i^{S,b_S}},\;\nabla_{e_i^{R,b_R}}$

        \[
        \nabla_{\theta_T}
        = \sum_{i\in\mathcal{B}}
        \nabla_{e_i^T}\,\frac{\partial e_i^T}{\partial \theta_T}
        \]

        \[
        \nabla_{\theta_S^{b}}
        = \sum_{i:\,b_S=b}
        \nabla_{e_i^{S,b}}\,\frac{\partial e_i^{S,b}}{\partial \theta_S^{b}}, 
        \qquad
        \nabla_{\theta_R^{b}}
        = \sum_{i:\,b_R=b}
        \nabla_{e_i^{R,b}}\,\frac{\partial e_i^{R,b}}{\partial \theta_R^{b}}
        \]

        \State $\theta_F \leftarrow \theta_F - \eta\,\nabla_{\theta_F}$
        \State $\theta_T \leftarrow \theta_T - \eta\,\nabla_{\theta_T}$

        \For{each bank $b$}
            \State $\theta_S^b \leftarrow \theta_S^b - \eta\,\nabla_{\theta_S^b}$
            \State $\theta_R^b \leftarrow \theta_R^b - \eta\,\nabla_{\theta_R^b}$
        \EndFor

    \EndFor

    \State Federated average of $\{\theta_S^b,\theta_R^b\}_{b\in\mathcal{B}}$

\EndFor

\end{algorithmic}
\end{algorithm}

\section{Experimental Setup}
This section describes the datasets, model architectures, and evaluation procedure used in the experiments. The proposed hybrid federated learning algorithm is mainly compared with the centralised model, where all data are available in one place, and additionally also with a local model trained only on the data held by the active (in this case, transaction) party. The goal is to evaluate how well the algorithm can learn when the data are distributed across parties in a hybrid partition.
\subsection{Datasets}
The experiments are conducted using synthetic datasets that reflect financial transaction behaviour.
\subsubsection*{AMLSim Dataset}
This dataset was generated using AMLSim \cite{altman2023realistic}, which is a multi-agent simulator that generates synthetic banking transactions together with predefined money-laundering behaviours. It produces transaction records and corresponding account profiles that mimic the structure of real financial activity. In our setting, the transaction-level attributes generated by AMLSim are assigned to the transaction party, while the associated account-level features are distributed between the participating banks according to the hybrid partition.
\subsubsection*{SWIFT Dataset}
This dataset was provided by SWIFT as part of the NSF-organized Privacy Enhancing Technologies Prize Challenge on Financial Crime Prevention \cite{nistUSUKPETs}. It contains transaction-level data and account-level features assigned to the transaction party and the banks in the same way as in the AMLSim dataset.

In both datasets, additional derived features were created for the three roles: the transaction party, the sender side, and the receiver side. These features were introduced to provide more informative inputs, capture behavioural aspects not directly available in the raw data, and maintain a consistent feature structure across the two datasets.
\begin{table}[h]
\footnotesize
\centering
\caption{Datasets used for the experiments}
\label{tab:dataset_summary}
\begin{tabular}{lcccc}
\hline
Dataset & Banks & Transactions & Positive:Negative Samples \\ 
\hline
AMLSim & 2 & 63330 & 70:30 \\
SWIFT  & 10 & 810563 & 903:1 \\
\hline
\end{tabular}
\end{table}

\begin{table*}
\centering
\scriptsize
\renewcommand{\arraystretch}{1.2}
\caption{Feature allocation across parties for AMLSim and SWIFT datasets}
\label{tab:hybrid_features}
\begin{tabular}{@{}p{1.2cm}@{}p{3.2cm}@{}p{4.2cm}@{}p{4.5cm}@{}p{4.5cm}@{}}
\toprule
\textbf{Dataset} & \textbf{Transaction Party Features} & \textbf{Bank: General Account Features} & \textbf{Sender-Specific Features} & \textbf{Receiver-Specific Features} \\
\midrule

\textbf{AMLSim} &
\begin{tabular}[t]{@{}l@{}} 
Transaction amount \\
Timestamp of transaction \\
Transaction type \\
Fraud label  \\
\textit{Identifiers:} \\transaction ID\\ sender ID\\ receiver ID
\end{tabular} &
\begin{tabular}[t]{@{}l@{}} 
Account type\\
Account status \\
Country of residence \\
Gender of account holder \\
Count of prior suspicious reports \\
Initial deposit amount \\
Age of account \\
\textit{Identifiers:} account ID, bank ID
\end{tabular} &
\begin{tabular}[t]{@{}l@{}} 
No. of outgoing transactions \\
Total amount sent by account \\
Average amount sent per transaction \\
\textit{Identifier:} account ID (sender role)
\end{tabular} &
\begin{tabular}[t]{@{}l@{}} 
No. of incoming transactions \\
Total amount received by account \\
Average amount received per transaction \\
\textit{Identifier:} account ID (receiver role)
\end{tabular} \\
\\\\

\textbf{SWIFT} &
\begin{tabular}[t]{@{}l@{}} 
Settlement amount \\
Transaction chain length \\
Settlement currency \\
Fraud label\\
\textit{Identifiers:} \\transaction ID\\ sender ID\\ receiver ID
\end{tabular} &
\begin{tabular}[t]{@{}l@{}} 
Account risk flags \\
New account indicator \\
Suspicious account indicator \\
\textit{Identifiers:} account ID, institution ID
\end{tabular} &
\begin{tabular}[t]{@{}l@{}} 
No. of transactions sent \\
Total amount sent by account \\
Average amount per sent transaction \\
Variation in sent transaction amounts \\
Velocity of outgoing transactions (past 24h) \\
\textit{Identifier:} account ID (sender role)
\end{tabular} &
\begin{tabular}[t]{@{}l@{}} 
No. of transactions received \\
Total amount received by account \\
Average amount per received transaction \\
Variation in received transaction amounts \\
Velocity of incoming transactions (past 24h) \\
\textit{Identifier:} account ID (receiver role)
\end{tabular} \\

\bottomrule
\end{tabular}
\end{table*}

\subsection{Model Architecture}
The architecture consists of four neural network models: a transaction encoder that transforms transaction-level attributes into a latent representation; a sender encoder and a receiver encoder, each responsible for extracting behavioural patterns from the account data held by the initiating and receiving banks, respectively; and a top model that combines these representations to perform binary fraud classification. All four models are implemented as multilayer perceptrons with ReLU activations and are jointly optimised to enable end-to-end learning across distributed data. Each encoder outputs a fixed-length embedding of dimension $P$, while the top model receives the concatenation of the three encoder outputs, forming a composite input of size $E = 3P$. This dimensionality is dataset-specific: for AMLSim, the embedding size is set to $P = 16$, resulting in a top model input of $E = 48$; for the SWIFT dataset, which involves a richer set of features and higher data complexity, the embedding size is increased to $P = 64$, yielding $E = 192$.

\subsection{Training Parameters}
Both models were trained using the Adam optimizer and a batch size of 128. For AMLSim, binary cross-entropy loss was used, while the SWIFT model uses focal loss \cite{lin2017focal}, which reduces the relative loss for well-classified examples and focuses learning on harder, misclassified cases. In the focal loss, $\alpha = 0.99$ adjusts the weight of positive samples, and $\gamma = 2.0$ controls the degree of down-weighting for easy examples. The learning rate was set to 0.00005 for AMLSim and 0.00001 for SWIFT. A dropout rate of 0.2 was applied in the SWIFT model to reduce overfitting. For AMLSim, the dataset was split into 70\% training, 15\% validation, and 15\% test sets. For SWIFT, the split was 75\% training and 25\% validation, with a separate test set of 500k transactions provided. Evaluation focused primarily on the area under the precision-recall curve (AUPRC), as it provides a more informative measure of performance in highly imbalanced classification tasks. In addition, precision, recall, and F1 score were reported to assess the quality of prediction at a specific threshold.
\section{Results}
We trained the model using the HybridFL algorithm and evaluated whether it could learn effectively from distributed data partitions. Experiments were carried out on the AMLSim and SWIFT datasets, which both contain imbalanced fraud-detection labels and reflect the hybrid structure considered in this work. To measure performance, we used the Area Under the Precision–Recall Curve (AUPRC) as the primary metric, since it is more informative than accuracy in imbalanced settings. In addition, we report precision, recall, and F1 score at a fixed threshold to show how the models behave when making discrete predictions. These threshold-based metrics complement AUPRC but are not the main evaluation target.
\begin{table}[h]
\centering
\footnotesize
\caption{Performance of the HybridFL algorithm on the AMLSim dataset. 
AUPRC reflects performance across all thresholds, while precision, recall, and F1 score are reported at a fixed threshold of 0.5.}
\label{tab:amlsim_results}
\begin{tabular}{lcccc}
\toprule
\textbf{Model} & \textbf{Precision} & \textbf{Recall} & \textbf{F1 Score} & \textbf{AUPRC} \\
\midrule
Centralised      & 0.88 & 0.77 & 0.82 & 0.91 \\
HybridFL             & 0.81 & 0.64 & 0.72 & 0.80 \\
Transaction Party Only & 0.39 & 0.50 & 0.44 & 0.53 \\
\bottomrule
\end{tabular}
\end{table}
\begin{table}[h]
\centering
\footnotesize
\caption{Performance of the HybridFL algorithm on the SWIFT dataset. 
AUPRC reflects performance across all thresholds, while precision, recall, and F1 score are reported at a fixed threshold of 0.5.}
\label{tab:swift_results}
\begin{tabular}{lcccc}
\toprule
\textbf{Model} & \textbf{Precision} & \textbf{Recall} & \textbf{F1 Score} & \textbf{AUPRC} \\
\midrule
Centralised      & 1.00 & 0.82 & 0.90 & 0.84 \\
HybridFL             & 0.92 & 0.75 & 0.83 & 0.78 \\
Transaction Party Only & 1.00 & 0.19 & 0.31 & 0.65 \\
\bottomrule
\end{tabular}
\end{table}

To understand how well HybridFL performs under hybrid data partitioning, we compared it with two reference models on both datasets. The first is a centralised model trained on all features from all parties in a single location, which serves as an upper bound on achievable performance. The second is a transaction-only model trained using only the features held by the transaction party. This provides a lower bound and shows how the model performs when account-level information from banks is not available. Comparing HybridFL with these two baselines allows us to assess (i) how close HybridFL comes to the centralised model and (ii) how much it improves over using only transaction-party features.

The results (Table \ref{tab:amlsim_results} \& \ref{tab:swift_results}) from both datasets show that the proposed HybridFL algorithm is able to learn effectively from distributed data. In AMLSim and SWIFT, its performance falls between the centralised model and the transaction-only baseline. This indicates that the model benefits from the information held at the banks, even though this information is never shared directly. One consistent observation is the clear improvement over the transaction-only model. On AMLSim, AUPRC increases from 0.53 to 0.80 (Table \ref{tab:amlsim_results}), and on SWIFT from 0.65 to 0.78 (Table \ref{tab:swift_results}). These gains show that the account-level features stored at the banks contain important behavioural signals that cannot be recovered from transaction-level features alone. The same pattern appears in the threshold-based metrics. The hybrid model achieves noticeably higher recall, meaning it identifies more minority-class fraud cases while maintaining reasonable precision.

The difference between the hybrid and centralised results is expected. In a centralised setting, all features are assumed to be stored in one place, allowing the model to learn relationships across the full feature space without any separation between transaction, sender, and receiver information. There is no need to generate separate embeddings or to coordinate gradient updates across multiple parties. This assumes that data can be freely combined, without privacy or confidentiality restrictions, and therefore represents an ideal upper bound. In the hybrid setting, the sender and receiver representations must be learned independently at each bank, and training relies on distributed gradients and periodic synchronisation. These constraints naturally limit how closely the hybrid model can match the centralised benchmark. What matters is that it still performs far better than the active party could achieve on its own, which was the main aim of the method. The two datasets highlight different levels of difficulty. AMLSim has a higher proportion of fraud cases and clearer behavioural patterns, which results in higher performance across all models. The SWIFT dataset is much more imbalanced, with a fraud ratio near 1:900, and presents a more challenging detection problem. This is reflected in a larger gap between the hybrid and centralised models. Even in this setting, the hybrid approach still provides a substantial improvement over the transaction-only baseline, showing that the additional account-level information remains valuable even under extreme imbalance. The threshold-based results (precision, recall, F1-score) in both further illustrate the limitations of relying only on transaction features. On SWIFT, the transaction-only model achieves high precision but very low recall, meaning it identifies almost none of the fraudulent transactions. This highlights the limited discriminatory power of transaction attributes in isolation. The hybrid approach increases recall considerably, demonstrating that the account-level embeddings contribute information that helps distinguish suspicious from legitimate behaviour.

Overall, the findings show that the proposed approach captures much of the predictive value available in the distributed features while preserving data locality. Although some performance is inevitably lost compared with the fully centralised case, the method offers a practical balance between model accuracy and confidentiality. It enables collaborative learning in settings where sharing raw account-level data is not feasible, while still achieving significantly better performance than the active party could obtain alone.
\section{Conclusion}
In this paper, we introduced a hybrid federated learning approach that works with data split both horizontally and vertically across participants. Our aim was to show that a model can still learn effectively when different parts of the data remain local. The results confirm that the hybrid model performs better than training on the active party alone, even though the data are distributed, and it comes reasonably close to a fully centralised model. As future work, we will examine the privacy properties of this approach. Because the method shares intermediate embeddings, we plan to add privacy-preserving mechanisms, such as differential privacy or secure aggregation, to limit what these embeddings might reveal. This will help us understand how much accuracy is lost when stricter privacy is applied and where useful trade-offs between privacy and performance can be made.

\bibliographystyle{IEEEtran}  
\bibliography{ref}     

\end{document}